\documentclass[10pt,letterpaper]{article}

\usepackage{main}
\usepackage{pslatex}
\usepackage{apacite}
\usepackage{hyperref}
\usepackage{natbib}
\usepackage{lineno}
\usepackage{amsfonts}
\usepackage{url}
\usepackage{graphicx} 
\usepackage{floatflt}
\usepackage{float}
\usepackage{algorithm}
\usepackage{algpseudocode}
\usepackage{amsmath}

\title{An explainable transformer circuit for compositional generalization}
\author{{\large \bf Cheng Tang\textsuperscript{1,2}} \\
  \AND {\large \bf Brenden Lake\textsuperscript{3}} \\
  \And {\large \bf Mehrdad Jazayeri\textsuperscript{1,2,4,*}}
  }

\begin{document}

\maketitle
\footnotetext[1]{Department of Brain and Cognitive Sciences, Massachusetts Institute of Technology, MA, USA}
\footnotetext[2]{McGovern Institute, Masachusetts Institute of Technology, MA, USA}
\footnotetext[3]{Center for Data Science, Department of Psychology, New York University, NY, USA}

\footnotetext[4]{Howard Hughes Medical Institute, MA, USA}

\section{Abstract}
{
\bf
Compositional generalization—the systematic combination of known components into novel structures—remains a core challenge in cognitive science and machine learning. Although transformer-based large language models can exhibit strong performance on certain compositional tasks, the underlying mechanisms driving these abilities remain opaque, calling into question their interpretability. In this work, we identify and mechanistically interpret the circuit responsible for compositional induction in a compact transformer. Using causal ablations, we validate the circuit and formalize its operation using a program-like description. We further demonstrate that this mechanistic understanding enables precise activation edits to steer the model’s behavior predictably. Our findings advance the understanding of complex behaviors in transformers and highlight such insights can provide a direct pathway for model control.
}
\begin{quote}
\small
\textbf{Keywords:} 
Transformer; Mechanistic Interpretability; Compositionality
\end{quote}

\section{Introduction}

Transformers, first introduced by \citet{Vaswani2017-cg}, excel at tasks requiring complex reasoning such as code synthesis~\citep{Chen2021-ys} and mathematical problem-solving~\citep{Hendrycks2020-ca}. This capability stems not merely from memorization, but from their ability to perform \emph{compositional generalization}—systematically combining learned primitives into novel structures via in-context learning (ICL)~\citep{Brown2020-ut,Lake2023-cp}. While humans inherently excel at such abstraction~\citep{Fodor1979-eg}, traditional neural architectures struggle with out-of-distribution (OOD) compositional tasks~\citep{Hupkes2019-qi,Lake2016-kh}. Understanding how neural systems accomplish compositionality has become a focus of both machine learning and cognitive science research.  

Mechanistic interpretability—a field dedicated to reverse-engineering neural networks into human-understandable algorithms—has begun unraveling these dynamics. Seminal work identified \emph{induction heads} as a critical component for ICL \citep{Elhage2021-vq, Olsson2022-hk}, enabling transformers to dynamically bind and retrieve contextual patterns rather than relying on shallow ``lazy'' heuristics like $n$-gram matching. However, prior works mostly focused on isolated mechanisms \citep{Wang2022-ha, Hanna2023-zy} or over-simplified models (e.g., attention-only transformer in \citet{Olsson2022-hk} , single layer transformer in \citet{Nanda2022-ds}), leaving interpretation of complex induction mechanisms in full-circuit transformers rarely explored. Furthermore, while studies have shown that hyper-parameters (e.g., number of attention layers) can causally affect model's compositional ability \citep{Sanford2024-ye,He2024-ws}, a microscopic inspection to the internal circuitry is still lacking.  

In this case study, we provide an end-to-end mechanistic interpretation of how a compact transformer solves a compositional induction task. We rigorously trace down the minimal circuit responsible for the model’s behavior and fully reverse-engineer the attention mechanism into human-readable pseudocode. We also bridge mechanistic interpretation and model control by showing that we can steer the model's behavior with activation edits guided by the circuit mechanism.

% Related Work section
\section{Related Work}
\label{appendix0}

\paragraph{Transformer circuit interpretation.}
Mechanistic interpretability of transformers began with analysis of simplified models, identifying attention heads as modular components that implement specific functions. In their seminal work, \citet{Elhage2021-vq} and \citet{Olsson2022-hk} introduced "induction heads" as critical components for in-context learning in small attention-only models. These heads perform pattern completion by attending to prior token sequences, forming the basis for later work on compositional generalization. Case studies have dissected transformer circuits for specific functions, such as the 'greater than' circuit \citep{Hanna2023-zy}, the 'docstring' circuit \citep{Heimersheim2023-ka}, the 'indirect object' circuit \citep{Wang2024-vp}, and the 'max of list' circuit \citep{Hofstatter2023-rv}. These case studies successfully reverse-engineered the transformer into the minimal-algorithm responsible for the target behavior.

To facilitate identification of relevant circuits, researchers have proposed circuit discovery methods such as logit lens \citep{nostalgebraist2020-du}, path patching \citep{Goldowsky-Dill2023-zr}, causal scrubbing \cite{LawrenceC2022-zy}.   For large-scale transformers, automated circuit discovery methods are also proposed \citep{Conmy2023-ej, Hsu2024-xo, Bhaskar2024-mn}. So far, transformer interpretability work still requires extensive human efforts in the loop for hypothesis generation and testing. We point to a review paper for a more comprehensive review \citep{Rai2024-yg}.

\paragraph{Compositional generalization in transformers.} In their study, \citet{Hupkes2019-qi} evaluated compositional generalization ability on different families of models, and found that transformer outperformed RNN and ConvNet in systematic generalization, i.e., recombination of known elements, but still uncomparable to human performance.  \citet{Zhang2024-yl} pointed out that transformers struggle with composing recursive structures. Recently, \citet{Lake2023-cp} showed that after being pre-trained with data generated by a 'meta-grammar', small transformers (less than 1 million parameters) can exhibit human-like compositional ability in novel in-context learning cases. This is in line with the success of commercial large language models (LLM) in solving complex out-of-distribution reasoning tasks \citep{Bubeck2023-ho,DeepSeek-AI2024-bs}, where compositional genralization is necessary. 

Several studies highlighted factors that facilitate transformer's compositional ability. \citet{Wang2024-mi} identified initialization scales as a critical factor in determining whether models rely on memorization or rule-based reasoning for compositional tasks. \citet{Zhang2025-id} revealed that low-complexity circuits enable out-of-distribution generalization by condensing primitive-level rules. \citep{Sanford2024-ye} identified logarithmic depth as a key constraint for transformers to emulate computations within a sequence. Here, we offer a complementary mechanistic understanding of how trasnformers perform compositional computations.

\section{Experimental Setup}

\begin{figure}[h]
\begin{center}
\includegraphics[width=0.35\textwidth]{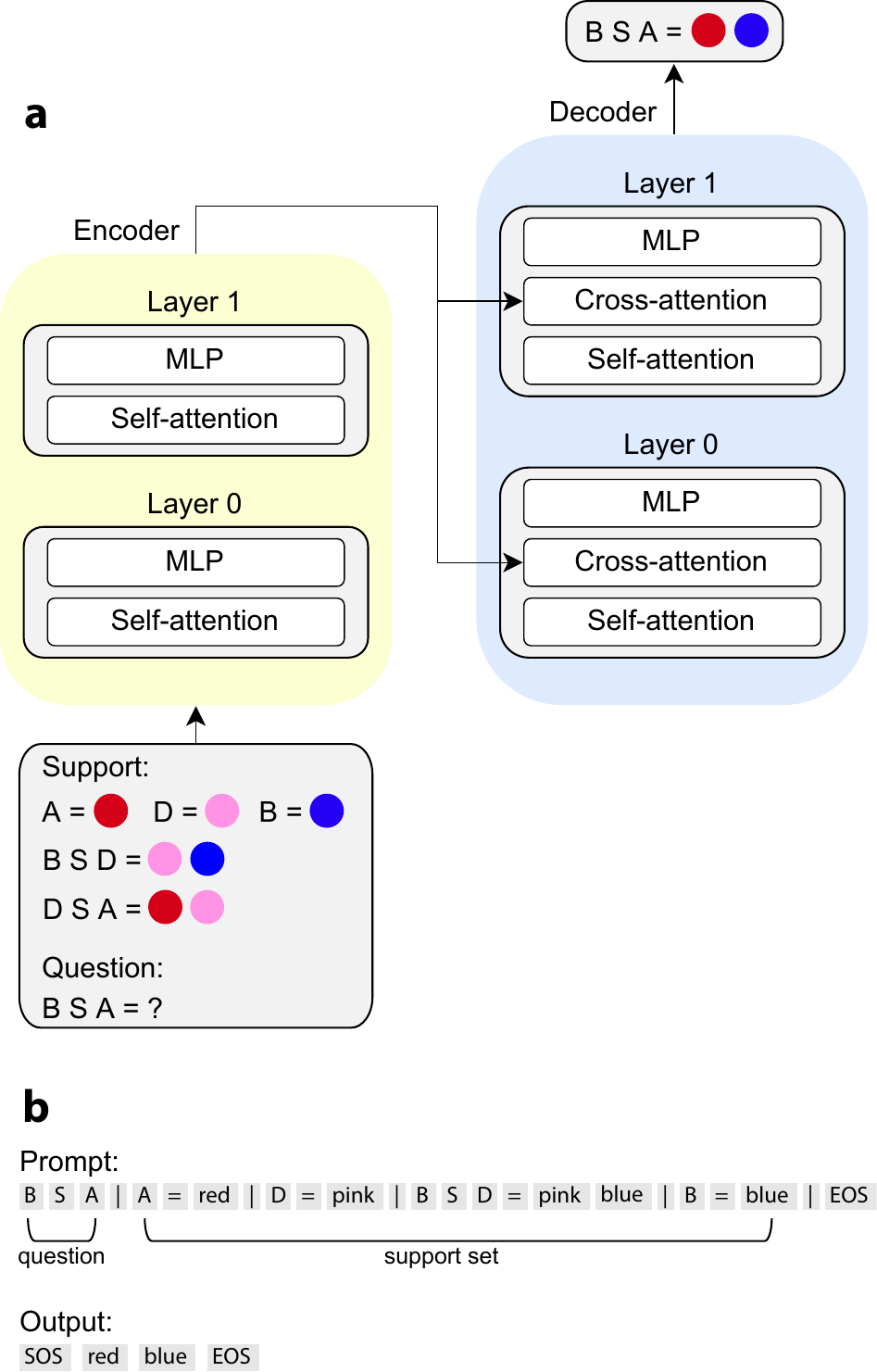}
\end{center}
\caption{(a) Schematic of the transfomer model and task. (b) The prompt and output format for the compositional induction task.}
\label{fig1}
\end{figure}

Our experimental setup involves a synthetic function composition task 
(Figure~\ref{fig1}) designed to probe \emph{compositional induction} in a compact Transformer. We outline the task structure, the Transformer basics (including attention mechanisms), and the training protocol.

\subsection{Task Structure}
\label{sec:task-structure}

Each episode consists of a \textbf{support set} and a \textbf{question} (Figure ~\ref{fig1}b): 

\begin{itemize}
    \item \textbf{Support Set:} Specifies (i) \emph{Primitives} as symbol-to-color mappings 
    (e.g., \texttt{A = red}, \texttt{D = pink}), and (ii) \emph{Functions} as symbolic operations 
    over these primitives (e.g., \texttt{A S D = pink red}, where \texttt{S} indicates swapping 
    adjacent symbols).
    \item \textbf{Question:} Presents a new composition of primitives and functions from the support-set. 
\end{itemize}

The model generates answers to the \textbf{question} as token sequences emitted from the decoder,  with a \texttt{SOS} (start of sentence) token as the first input to the decoder and an \texttt{EOS} (end of sentence) marking the end of the emission. The model operates strictly via in-context learning—weights remain frozen during inference, and test episodes are disjoint from training data. The model must infer latent variable bindings (primitives and functions) from the \textbf{support set} and dynamically compose these bindings to solve the novel \textbf{question}.

\subsection{Model}

\subsubsection{Transformer Basics}
\label{sec:transformer-basics}

Our transformer uses an encoder-decoder architecture that involves two types of attentions:
\begin{itemize}
    \item \textbf{Self-Attention:} Captures within-sequence interactions. 
    % The token embedding matrix $X \in \{R}^{n_{\text{input}} \times d_{\text{model}}}$ is projected into Queries, Keys, and Values:
    The token embedding matrix \[X \in \mathbb{R}^{n_{\text{input}} \times d_{\text{model}}}\] is projected into Queries, Keys, and Values:
    \[
      Q = X W_Q, \quad
      K = X W_K, \quad
      V = X W_V,
    \]
    where $W_Q, W_K, W_V \in \mathbb{R}^{d_{\text{model}} \times d_{\text{head}}}$ are learnable weight matrices.

    \item \textbf{Cross-Attention:} Enables the decoder to attend to encoder outputs. Here,
    the Queries ($Q$) come from the \emph{decoder} tokens, while the Keys ($K$) and Values ($V$) 
    come from the \emph{encoder} tokens.
\end{itemize}

The attention mechanism operates through two separate circuits on embedding $X \in \mathbb{R}^{n_{\text{input}} \times d_{\text{model}}}$ for each attention head:  
\begin{itemize}
    \item \textbf{QK Circuit} ($W_QW_K^\top$): Determines from \textit{where} information flows to each token by computing attention scores between token pairs, with higher scores indicate stronger token-to-token relationships:
    \[
    \text{Attention}(Q,K) = \text{softmax}\left(\frac{X_QW_Q (X_KW_K)^\top}{\sqrt{d_{\text{head}}}}\right) \in \mathbb{R}^{n_{\text{query}} \times n_{\text{key}}},
    \]
    where \textit{softmax} is applied along the dimension for \textit{Key} and independently for each head.

    \item \textbf{OV Circuit} ($W_VW_O$): Controls \textit{what} information gets written to each token position. Combined with the \textbf{QK Circuit}, this produces the output of the attention head:
    \[
    Z = \text{Attention}(Q,K)  X_VW_VW_O \in \mathbb{R}^{n_{\text{query}} \times d_{\text{model}}},
    \]
    where $W_O \in \mathbb{R}^{d_{\text{head}} \times d_{\text{model}}}$ is learnable weight.
\end{itemize}
Our analysis focuses on how these circuits in attention heads together implement the compositional induction algorithm.

\subsubsection{Model Training}
\label{sec:architecture}

We adopt an encoder-decoder Transformer with \textbf{2 layers} in the encoder and \textbf{2 layers} 
in the decoder (Figure \ref{fig1}a) with each layer containing \textbf{8 attention heads}. Further model details appear in 
the Appendix.

\label{sec:dataset-training}

For each episode, we randomly generate:
\begin{itemize}
    \item \textbf{Primitive Assignments}: A mapping from symbol tokens (e.g., \texttt{A}, \texttt{B})
    to color tokens (e.g., \texttt{red}, \texttt{pink}).
    \item \textbf{Function Definitions}: Symbolic transformations by randomly sampling primitive arguments to a function to produce color sequences (e.g., \texttt{A S B} might be expanded 
    into a sequence \texttt{[A][B][A][A][B]}, maximum length=5).
\end{itemize}

We train on 10{,}000 such episodes for 50 epochs and evaluate on 2{,}000 test (held-out) episodes. The model
achieves \textbf{98\%} accuracy on this test set, indicating strong compositional induction
capabilities. In the test set, primitive assignments and function definitions are conjunctively different from those in the training set (i.e., some primitives or some functions might be in the training set, but not the whole combination of them), preventing a memorization strategy. Please refer to the Appendix for additional details.

\section{Results}
\label{sec:results}

 First, we give an intuitive overview of the 
\emph{effective algorithm} the model appears to implement. Next, we 
describe our \emph{circuit discovery} procedure, where we use causal 
methods to pinpoint the exact attention heads responsible for compositional induction. Finally, we validate this mechanism by applying targeted perturbations 
that predictably alter the model’s behavior.

\subsection{The Effective Algorithm}

\textbf{General Solution.} We first provide a general solution to this type of compositional problem in a python-like pseudocode for intuitive understanding (Algorithm \ref{alg1}). We use $1$-indexing (count from 1) for tokens throughout.

\begin{algorithm}[h!]
\caption{Pseudocode solving the function \& primitive composition problem}
\label{alg1}
    
\textbf{\# Define the question and symbol-color pairs (by Question-Broadcast and Primitive-Pairing Heads)}\\
$\textit{question} \gets [\,\texttt{s}_1,\;\texttt{func},\;\texttt{s}_2\,]$  \# definition \\
$\textit{symbol\_to\_color} \gets \{\, \texttt{s}_i : \texttt{c}_i \mid i=1,\dots,n \}$ \# definition \\
$\textit{color\_to\_symbol} \gets \{\, \texttt{c}_i : \texttt{s}_i \mid i=1,\dots,n \}$ \# reverse definition \\

\textbf{\# 
% Parse the function’s transformation 
Define the function; Convert the function into a relational structure between the input and output 
(by Primitive- and Function-Retrieval Heads) }\\
$\textit{func\_LHS} \gets [\texttt{s}_3\ \texttt{func}\ \texttt{s}_4]$  \# define function arguments \\
$\textit{func\_RHS} \gets [\texttt{c}_3 \ \texttt{c}_3 \ \texttt{c}_4 \ \texttt{c}_4 \ \texttt{c}_3]$  \# define function outputs \\
$\textit{symbol\_to\_idx} \gets \{\, \texttt{s}_3 \texttt{:} \texttt{idx}_1 \; , \;   \texttt{s}_4 \texttt{:} \texttt{idx}_3\}$  \# convert argument symbols to their index in array  \\

$\textit{idx\_seq} \;\gets\; [\,]$

\textbf{for} {$\textit{color}$ \textbf{in} $\textit{func\_RHS}$} \textbf{do}\\
\phantom{for } $\textit{symbol}\gets \textit{color\_to\_symbol}[\textit{color}]$ \\
\phantom{for } $\textit{idx} \gets \textit{symbol\_to\_idx}[\textit{symbol}]$\\
\phantom{for } $\textit{idx\_seq}.\mathrm{append}(\textit{idx}) $

% Extract the function (e.g., ${\texttt{s}_3\ \texttt{func}\ \texttt{s}_4 = \texttt{c}_3\ \texttt{c}_3\ \texttt{c}_4\ \texttt{c}_4\ \texttt{c}_3}$).
% Map each color on the RHS back to the symbol's relative-index-on-LHS 
% (\(\texttt{s}_3 = \mathit{idx}_1,\ \texttt{s}_4 = \mathit{idx}_3\)) , yielding:

\#  $\textit{idx\_seq} \;= [\,\texttt{idx}_1,\;\texttt{idx}_1,\;\texttt{idx}_3,\;\texttt{idx}_3,\;\texttt{idx}_1\,]$ in this case \\

\textbf{\# 
% Compose a new output from the question 
Compose the output following the function's relational structure
(by RHS-Scanner and Output Heads)}

$\textit{output} \gets [\,]$ 

\textbf{for} {$\textit{idx}$ \textbf{in} $\textit{idx\_seq}$} \textbf{do} \\
\phantom{for } $\textit{symbol} \gets \textit{question}[\textit{idx}]$\\
\phantom{for } $\textit{color} \gets \textit{symbol\_to\_color}[\textit{symbol}]$\\
\phantom{for } $\textit{output}.\mathrm{append}(\textit{color}) $ \\
     
\Return $\textit{output}$ 
\end{algorithm}

\noindent
\textbf{Transformer Solution.} Next, we describe the actual implementation of the algorithm with attention operations in Figure \ref{fig_algo} through a guidance episode. 

\noindent
\subsubsection{Step 1 (Figure \ref{fig_algo}a; Question-Broadcast Head).} Primitive input tokens in the support (e.g., \texttt{A}) attend to the same primitive tokens in the question (\texttt{A}), inheriting the latter's index-in-question (\texttt{3rd}). The step is detailed in Figure \ref{fig5}b. 

\subsubsection{Step 2 (Figure \ref{fig_algo}b; Primitive-Pairing Head).} Color tokens (\texttt{red}) attend to their associated primitive tokens (\texttt{A}), inheriting the latter's index-in-question (\texttt{3rd}). The step is detailed in Figure \ref{fig5}a. 

\subsubsection{Step 3 (Figure \ref{fig_algo}c; Primitive- and Function-Retrieval Heads).} Color tokens on the function RHS (\texttt{pink}) attend to their associated primitive tokens on the function Left Hand Side (LHS) (\texttt{D}), inheriting the latter's relative-index-on-LHS (\texttt{3rd}). The step is detailed in Figure \ref{fig9}. 

\subsubsection{Step 4 (Figure \ref{fig_algo}d; RHS-Scanner Head).} The \texttt{1st} token in the Decoder (\texttt{SOS}) attend to the \texttt{1st} tokens on the function Right Hand Side (RHS) (\texttt{pink}), inheriting the latter's former-inherited relative-index-on-LHS (\texttt{3rd}). The step is detailed in Figure \ref{fig8}.  

\begin{figure}[h!]
\begin{center}
%\fbox{CCN figure}
\includegraphics[width=0.42\textwidth]{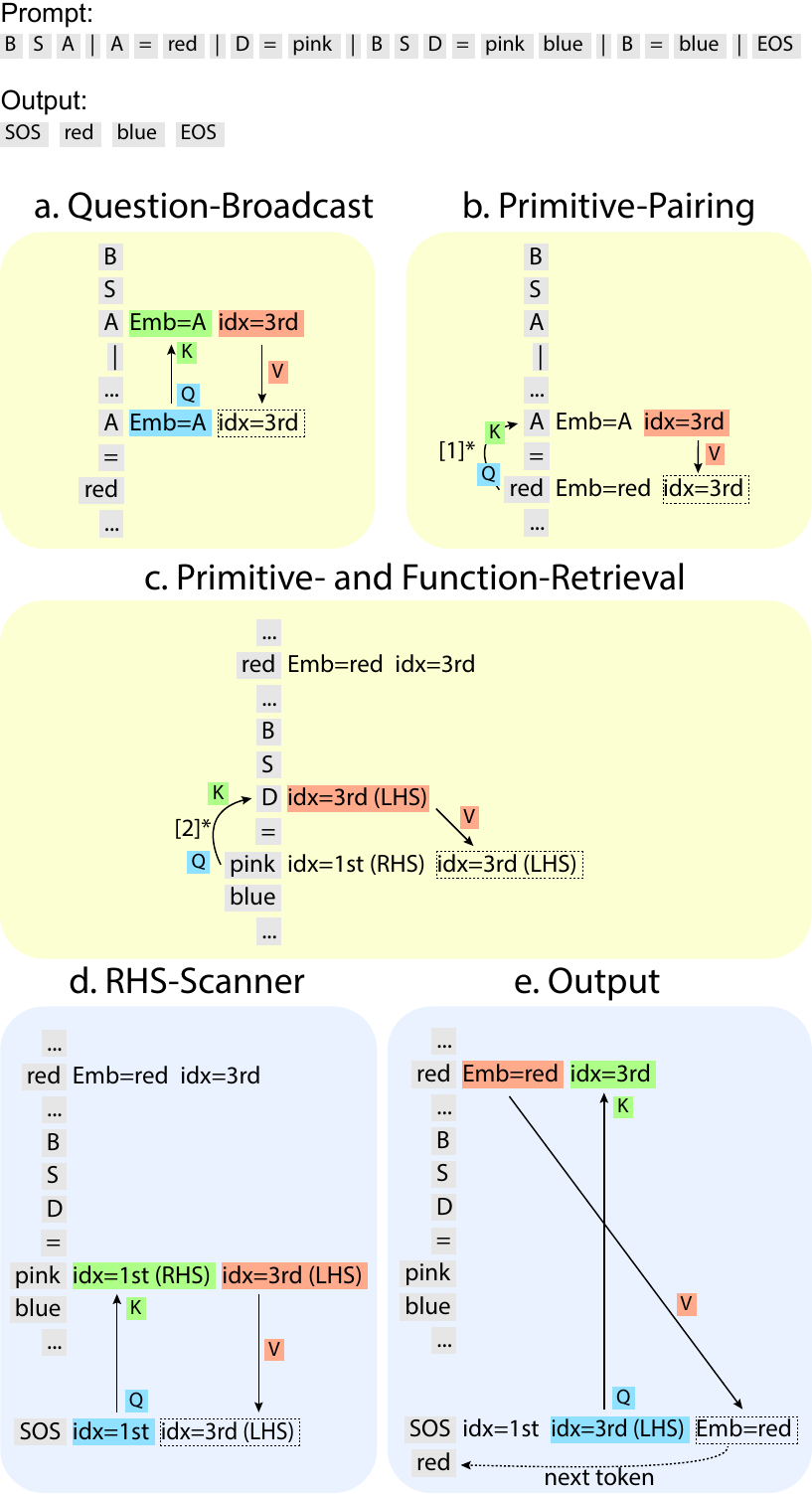}
\end{center}
\caption{
Summary of circuit for compositional generalization. Top, the example episode's input and output. For a-e, the yellow boxes indicate self-attention heads and the blue boxes indicate cross-attention heads. Titles refer to the functional attention heads that execute the steps (discussed in detail later).  We unfold all relevant information superimposed in tokens' embeddings and highlight their roles in attention operations. $[1]^*$, the $QK$ alignment discussed in Primitive-Pairing Head section. $[2]^*$, the $QK$ alignment discussed in Primitive-Retrieval Head section.
}
\label{fig_algo}
\end{figure}

\subsubsection{Step 5 (Figure \ref{fig_algo}e; Output Head).} \texttt{SOS} token (with inherited relative-index-on-LHS=\texttt{3rd}) attends to color tokens (\texttt{red}) with the same index-in-question (\texttt{3rd}), inheriting the latter's token identity (\texttt{red}), and generate the next prediction (\texttt{red}). The step is detailed in Figure \ref{fig2}. Then the \texttt{2nd} token in the Decoder (\texttt{red}) starts over from \textbf{Step 4} until completion of function RHS.

\subsection{Circuit Discovery}
\label{section4.2}
Nomenclature: for attention heads, \textit{Enc-self-0.5} stands for \textit{Encoder, self-attention, layer 0, head 5}; similarly, \textit{Dec-cross-1.5} stands for \textit{Decoder, cross-attention, layer 1, head 5}.

\begin{figure}[h!]
\begin{center}
\includegraphics[width=0.45\textwidth]{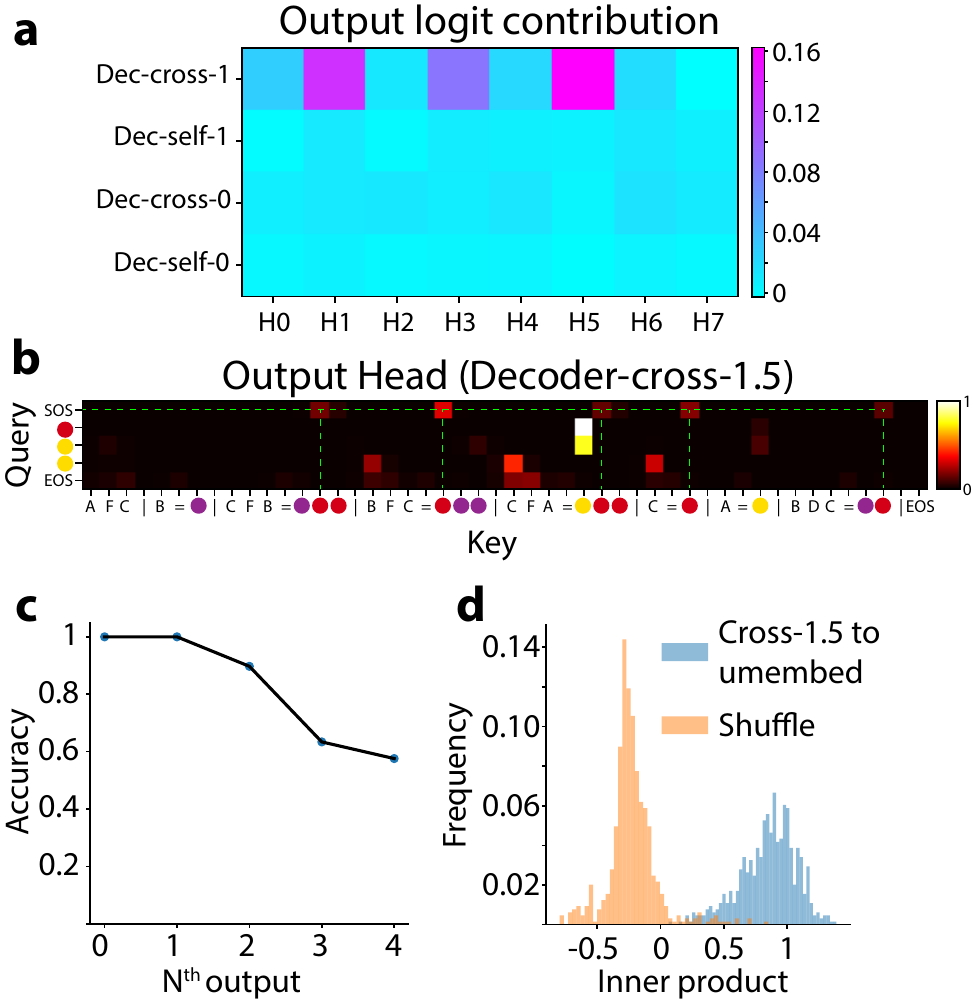}
\end{center}

\caption{(a) Logit contributions of each decoder head to the logits of correct tokens. (b) 
Attention pattern of Dec-cross-1.5. (c) For Dec-cross-1.5, the 
percentage of attention focused on the next predicted token. (d) For Dec-cross-1.5, alignment (inner product) between 
its $OV$ output (e.g., $x_{red}W_vW_o$) and the corresponding unembedding vector 
(e.g., \(\mathrm{Unemb}_{red}\)). We estimated the null distribution by randomly sampling unembedding vectors.
}
\label{fig2}
\end{figure}

\subsubsection{Output Head (Dec-cross-1.5; Figure \ref{fig2}b)}
\label{section:output-head-identification}

We discovered the model’s circuit backwards from the unembedding layer using \textit{logit attribution} \citep{nostalgebraist2020-du}, which measures each decoder attention head’s linear contribution to the final token logits 
(adjusted by the decoder’s output \textit{LayerNorm}). We identified 
\textbf{Dec-cross-1.5} (decoder cross attention layer 1 head 5) as the primary contributor (Figure~\ref{fig2}a).

Dec-cross-1.5’s $Q$ tokens always attend to the $K$ tokens from the Encoder that are the \emph{next} predicted ones. For example, in Figure~\ref{fig2}b, the \texttt{SOS} token attends to instances of \texttt{red} in the support set, which is indeed the correct next output prediction. 
This attention accuracy (i.e., max-attended token being the next-emitted token) of Dec-cross-1.5 remains above 90\% for the first three tokens in the responses across all test episodes (Figure \ref{fig2}c), with Dec-cross-1.1 and -1.3 partially compensating beyond that point. 

These observations suggest that Dec-cross-1.5’s $OV$ circuit feeds token identities  directly to the decoder unembedding layer (output layer). Specifically, we observe 
that the output of the $OV$ circuit, $XW_vW_o$, align closely (strong inner product) with the unembedding vectors of the corresponding tokens ({Figure \ref{fig2}d}).
Hence, we designate Dec-cross-1.5 as the \textit{Output Head} (while Dec-cross-1.1 and -1.3 perform similar but less dominant roles) (Algorithm Step 3).

Next, we show how the Output Head identifies the correct token through $QK$ interactions.

\subsubsection{The K-Circuit to the Output Head}
\label{section:k-circuit}

We first determine which encoder heads critically feed into the Output Head's $K$. To do 
this, we performed \textit{path-patching} \citep{Wang2022-ha} by ablating all but one single encoder head and then 
measuring how much of Output Head’s $QK$ behavior (\emph{i.e.}, attention accuracy) remained. During these experiments, Output Head’s $Q$ were 
\textit{frozen} using clean-run activations.  Here we report patching results with mean-ablation (qualitative similar to random-sample ablation) (details in Appendix).

\begin{figure}[h!]
\begin{center}
\includegraphics[width=0.32\textwidth]{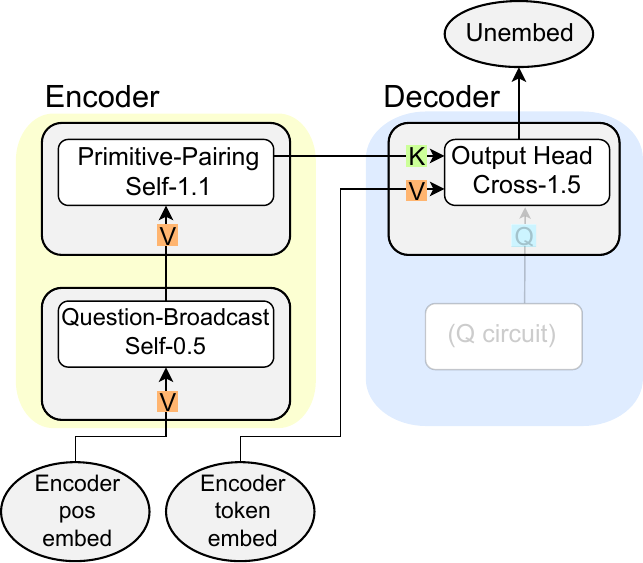}
\end{center}
\caption{Enc-self-1.1 and Enc-self-0.5 serve as the main contributors of the $K$-circuit for 
the Output Head.The $K$-circuit encodes primitive symbols' index-in-question.}
\label{fig4}
\end{figure}

Through this process, we identified \textbf{Enc-self-1.1} and \textbf{Enc-self-0.5} as the 
primary contributors to Output Head’s $K$, acting in a sequential chain (Figure~\ref{fig4}). Next, we show how they sequentially encode symbols' index-in-question critical for the $QK$ alignment. 

\begin{figure}[h!]

\begin{center}
\includegraphics[width=0.43\textwidth]{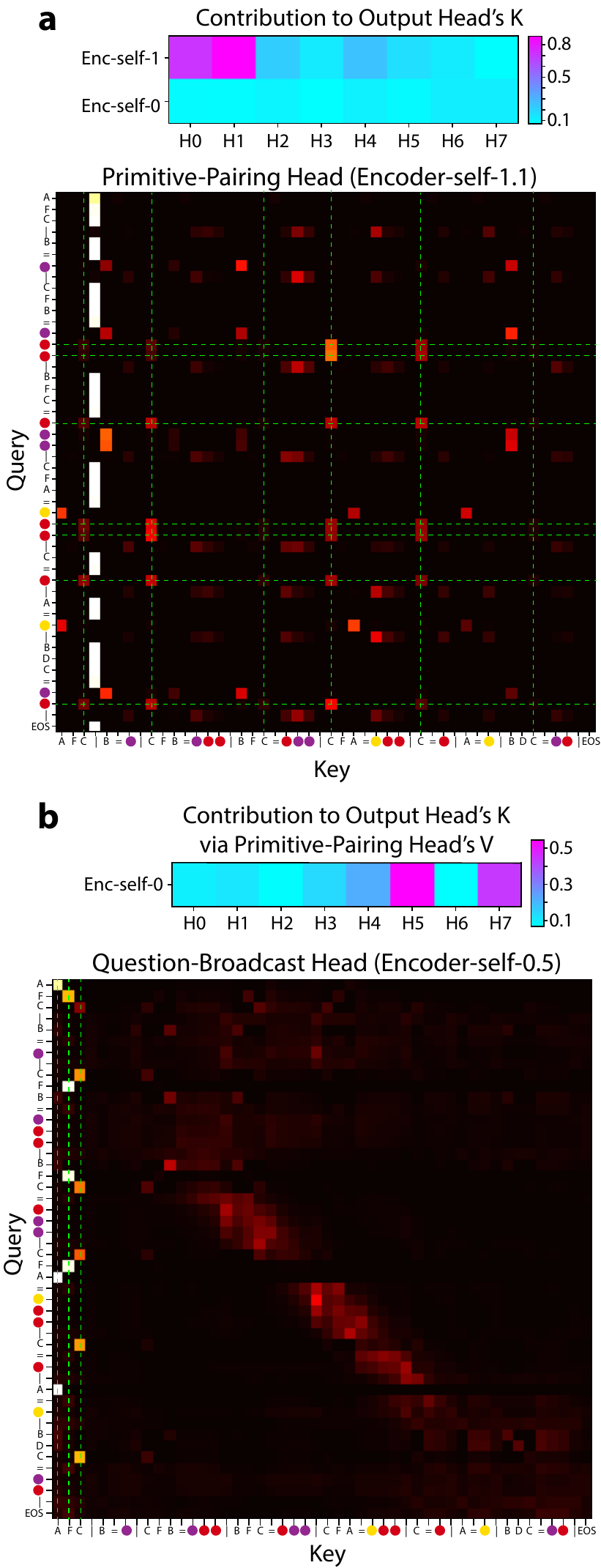}
\end{center}
\caption{(a) Top, contributions to Output Head’s performance (percentage of attention on the correct next 
token) via $K$. Bottom, attention pattern of Enc-self-1.1. (b) Top, contributions to the Output Head’s performance through the Primitive-Pairing Head's $V$. Bottom, attention pattern of Enc-self-0.5.
}
\label{fig5}

\end{figure}

\paragraph{Primitive-Pairing Head (Enc-self-1.1; Figure \ref{fig5}a)}
This head exhibits a distinct attention pattern that pairs each color token with its associated primitive symbol token (e.g., in the support set, all instances of \texttt{red} attend to \texttt{C}). In other words, Enc-self-1.1 relays information (described below, as computed by e.g., \textbf{Enc-self-0.5}) from the primitive symbols to their corresponding color tokens via its $QK$ circuit. Hence, we call Enc-self-1.1 the \textit{Primitive-Pairing Head}.

To investigate which upstream heads feed into the $OV$ circuit of the Primitive-Pairing Head, we applied a
sequential variant of path-patching, isolating the chain:

\begin{center}
Upstream heads (e.g.\ Enc-self-0.5)
$\;\longrightarrow\;$ \\
\text{Primitive-Pairing Head ($V$)}
$\;\longrightarrow\; $\\
\text{Output Head ($K$)},
\end{center}
while mean-ablating all other direct paths to Output Head’s $K$. We identified \textbf{Enc-self-0.5} as an important node (Figure \ref{fig5}b).

\paragraph{Question-Broadcast Head (Enc-self-0.5; Figure \ref{fig5}b)} All input symbol in the support set attend to their copies in the input question. In other words, Enc-self-0.5 broadcasts question-related information (including token identity and position) across symbols in the support-set (henceforth the Question-Broadcast Head). We hypothesize that the primitive symbols' index-in-question is the critical information passed from the Question-Broadcast Head's $Z$ through the Primitive-Pairing Head's $Z$ and lastly into the Output Head’s $K$.

% \begin{SCfigure}[1][h]  % [0.5] = width relative to text, [h] = placement
% \centering
% \includegraphics[width=0.7\textwidth]{ICLR 2025 Template/figures/figure5.pdf}
% \caption{(a) Top, contributions to Cross-1.5’s performance (percentage of attention on the correct next 
% token) via Key values. Bottom, attention pattern of Enc-self-1.1. (b) Top, contributions to Cross-1.5’s performance through Enc-self-1.1 Value. Bottom, attention pattern of Enc-self-0.5.
% }
% \label{fig5}
% \end{SCfigure}

\begin{figure}[h!]
\begin{center}
\includegraphics[width=0.38\textwidth]{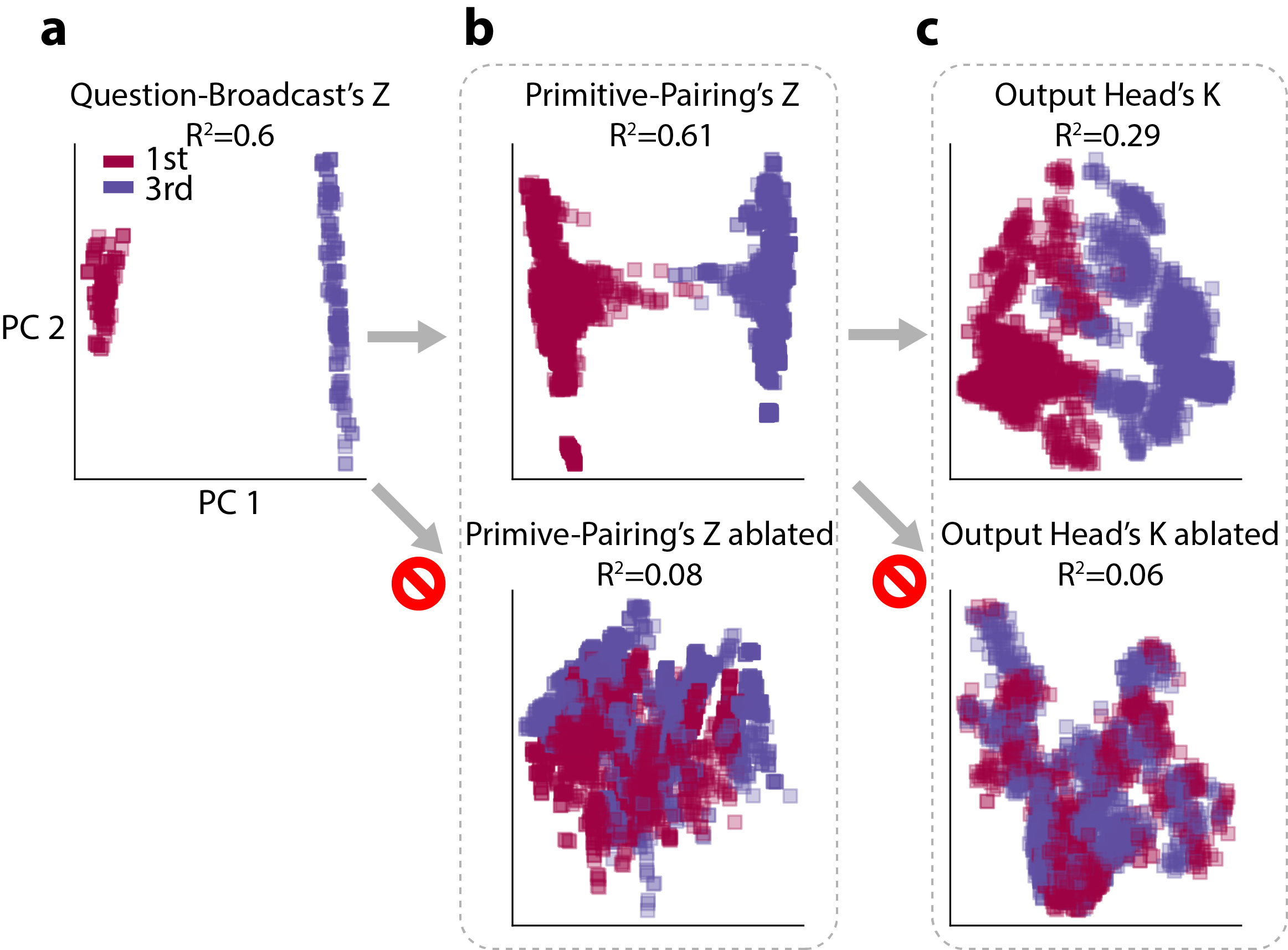}
\end{center}
\caption{Principal Components Analysis (PCA) of token embeddings, colored by their associated index-in-question. Concretely, for a prompt like '\texttt{B S A | A=red | B=blue | ...}', in (a), points are the $Z$ of '\texttt{A}' and '\texttt{B}' in the support (\texttt{A} labeled \texttt{3rd, B} labeled \texttt{1st}); in (b), points are the $Z$ of '\texttt{red}' and '\texttt{blue}' in the support (\texttt{red} labeled \texttt{3rd, blue} labeled \texttt{1st}); in (c), points are the $K$ of '\texttt{red}' and '\texttt{blue}' in the support (\texttt{red} labeled \texttt{3rd, blue} labeled \texttt{1st}). The distinct clusters suggest strong index information. $R^2$ score quantifies the percentage of total variance explained by the index identity.}
\label{fig6}
\end{figure}

\noindent
\textbf{Index-In-Question Tracing} 
To validate this hypothesis, we examined the Question-Broadcast Head's $Z$ for each 
\emph{primitive-symbol} token. We reduced these outputs to two principal components and 
colored each point by its index-in-question. As illustrated in Figure~\ref{fig6}a, the Question-Broadcast Head's $Z$ exhibit clear clustering, indicating that the index-in-question is robustly encoded at this stage (quantified by the $R^2$ score, i.e., the amount of variance explained by index identity, details in Appendix).
We further confirmed that the Primitive-Pairing Head's $Z$ preserves index-in-question (Figure \ref{fig6}b) and that the resulting Output Head's $K$ also reflect the same clustering (Figure \ref{fig6}c).

\textbf{Causal Ablation} Finally, we verified that this circuit indeed causally propagates index-in-question. Ablating the Question-Broadcast Head's $Z$ (together with the similarly functioning Enc-self-0.7) obliterates the clustering in the Primitive-Pairing Head's $Z$; ablating the Primitive-Pairing Head's $Z$ (together with similarly functioning Enc-self-1.0) disrupts the clustering in the Output Head's $K$ (Figure~\ref{fig6}). We therefore conclude that the Question-Broadcast Head, the Primitive-Pairing Head and heads with similar functions form a crucial $K$-circuit pathway, passing index-in-question information from primitive tokens to their associated color tokens in the Output Head's $K$.

\subsubsection{The Q-Circuit to the Output Head}
\label{section:q-circuit}
\begin{figure}[h!]
\centering
\includegraphics[width=0.4\textwidth]{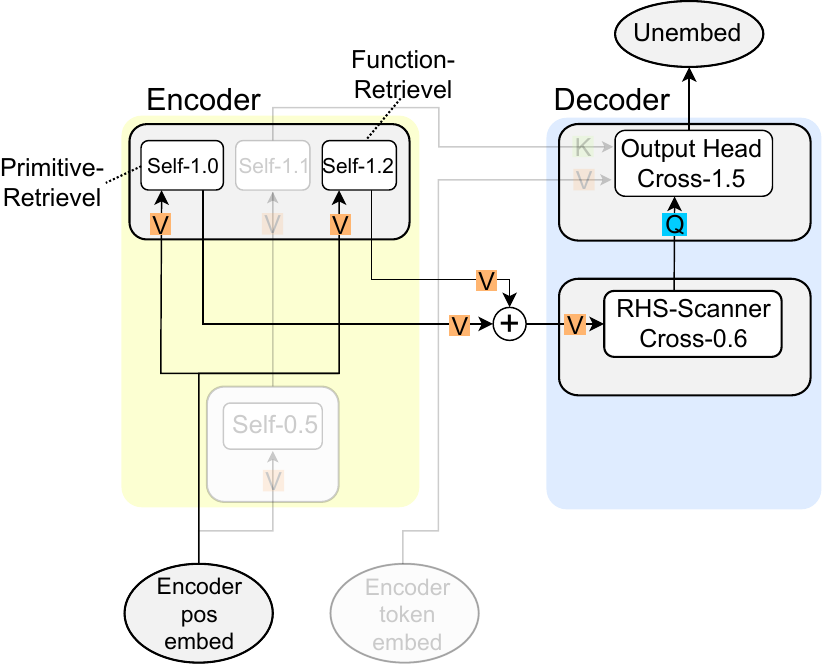}
\caption{Schematic of the $Q$-circuit. The Output Head inherits its $Q$ from Dec-cross-0.6, which
aggregates positional information passed from Enc-self-1.0 and Enc-self-1.2. The $Q$-circuit encodes primitive symbols' relative-index-on-LHS.}
\label{fig7}
\end{figure}

Having established the role of the $K$-circuit, we next investigate where its $Q$ originates. We again relied on \emph{sequential
path-patching} to pinpoint which decoder heads ultimately provide the Output Head's $Q$. We identified \textbf{Dec-cross-0.6} as the main conduit for the $Q$ values of the Output Head. Enc-self-1.0 and -1.2 supply positional embeddings that enable the decoder to track primitive symbol's relative-index-on-LHS, thereby completing the $QK$ alignment for correct predictions (Figure \ref{fig7}).

\begin{figure}[h!]
\begin{center}
\includegraphics[width=0.45\textwidth]{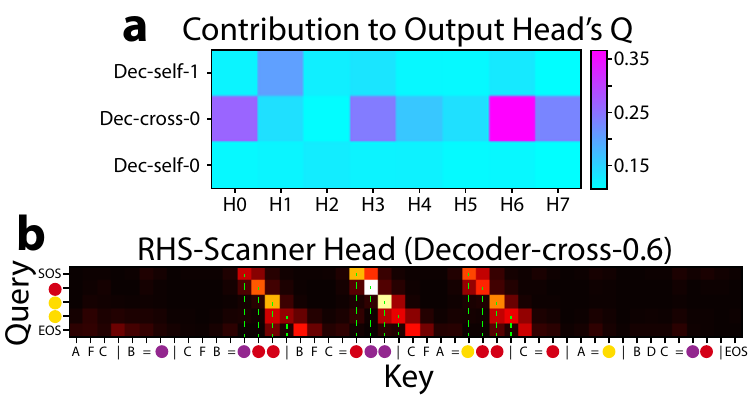}
\end{center}
\caption{(a) Contribution to Output Head's performance via $Q$.(b) Attention pattern of Dec-cross-0.6. }
\label{fig8}
\end{figure}

\paragraph{RHS-Scanner Head (Dec-cross-0.6; Figure \ref{fig8}b)}
We identify \textbf{Dec-cross-0.6} as the dominant contributor to the the Output Head's $Q$ (Figure \ref{fig8}a). 
Analyzing Dec-cross-0.6’s attention patterns reveals that each $Q$ token (from Decoder in the cross-attention) sequentially attends to the color tokens (in the support set) on the function’s RHS (Figure \ref{fig8}b). For example, the first Decoder token (\texttt{SOS}) attends to the first RHS tokens (\texttt{purple}, \texttt{red}, 
\texttt{yellow}), and the second query token (\texttt{red}) attends to the second RHS tokens 
(\texttt{red}, \texttt{purple}, \texttt{red}), and so on. This iterative scanning mechanism 
enables the decoder to reconstruct the transformation defined by the function. Hence we call Dec-cross-0.6 the RHS-Scanner Head.

\begin{figure}[h!]
\begin{center}
\includegraphics[width=0.42\textwidth]{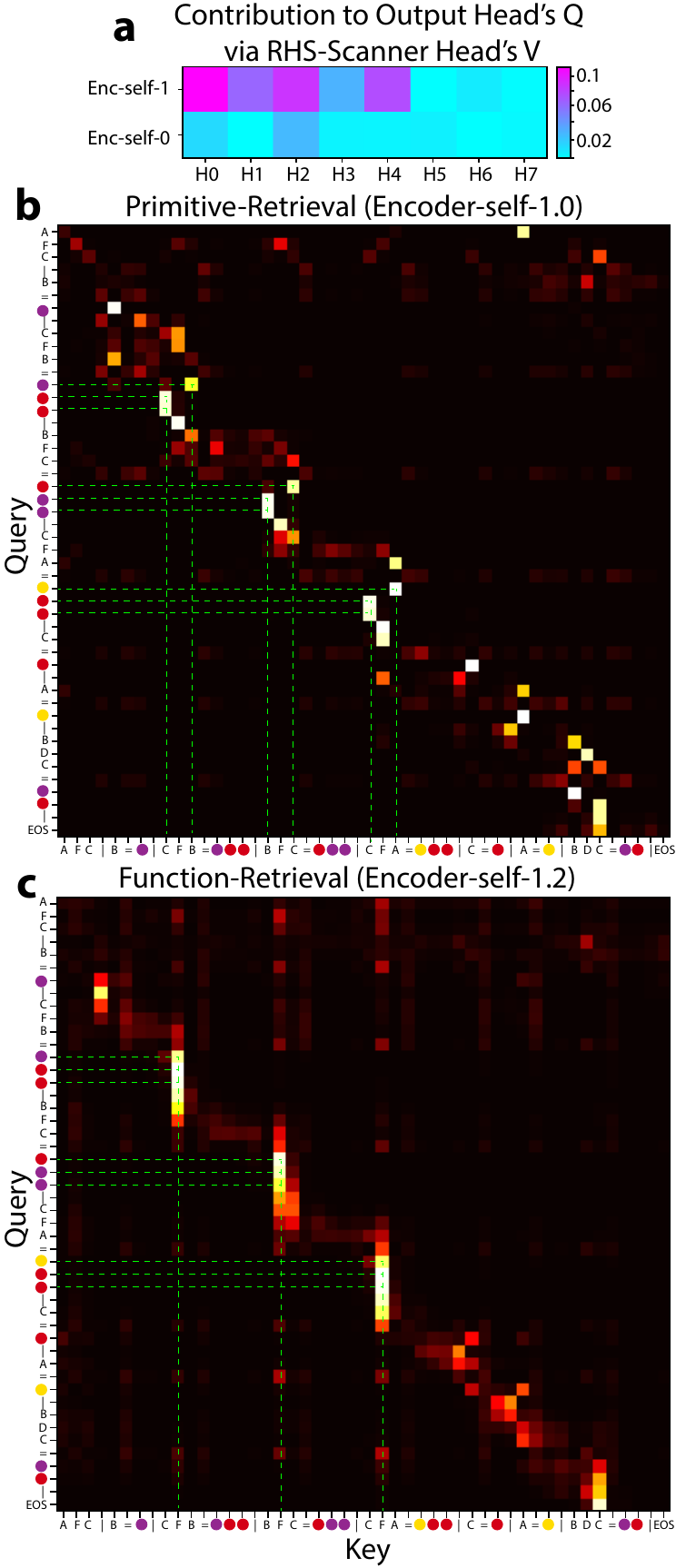}
\end{center}
\caption{(a) Contribution to Output Head's performance via $Q$. (b) Contribution to Output Head's performance via the RHS-Scanner's $V$. (c) Attention pattern of Dec-cross-0.6. (d) and (e) Attention 
patterns of Enc-self-1.0 and Enc-self-1.2.}
\label{fig9}
\end{figure}

\paragraph{Primitive-Retrival Head (Enc-self-1.0; Figure \ref{fig9}b) and Function-Retrival Head (Enc-self-1.2; Figure \ref{fig9}c)}
Next, we looked for critical encoder heads that feeds to the RHS-Scanner Head and finally contributes to the Output Head's $Q$. Unlike the $K$-circuit discovery, where “keep-only-one-head” ablations is sufficient, multiple heads appears to contribute partial but complementary information. To isolate their roles, we measured 
drops in the output head’s accuracy when ablating each encoder head individually while keeping the 
others intact (the ``ablate-only-one-head'' approach, more discussion in Appendix).

This analysis highlighted \textbf{Enc-self-1.0} and \textbf{Enc-self-1.2} as critical (Figure \ref{fig9}a). In 
Enc-self-1.0, within the support set, each color token on the RHS attends back to its corresponding symbol on the LHS, inheriting that symbol’s token and positional embedding (henceforth the Primitive-Retrieval Head) (Fig. \ref{fig9}b). Meanwhile, Enc-self-1.2 is similar, such that each color token on the RHS attends back to its function symbol on the LHS, passing that token and positional embedding on to the color token (henceforth the Function-Retrieval Head) (Fig. \ref{fig9}c). 

Why do the color tokens on the RHS attend back to both kinds of information on the LHS? We reason that if a color token on the RHS were to encode it's primitive symbol's relative-index-on-LHS: for example, in '\texttt{...| D=pink | A S D=pink red |...}', \texttt{pink} were to encode \texttt{3rd} inherited from \texttt{D} (\texttt{D} is \texttt{3rd} in '\texttt{A S D}'), the \textit{absolute} position of \texttt{D} must be compared with the \textit{absolute} position of the \texttt{S} to yield a \textit{relative} position. Now that with the Primitive- and Function-Retrievel Heads, each RHS color token carries two positional references: (1) the associated LHS primitive, and (2) the function symbol, we hypothesize that by comparing these references, the model can infer the primitive symbols' relative-index-on-LHS for each of the associated color tokens on the RHS.

\begin{figure}[h!]
\begin{center}
\includegraphics[width=0.46\textwidth]{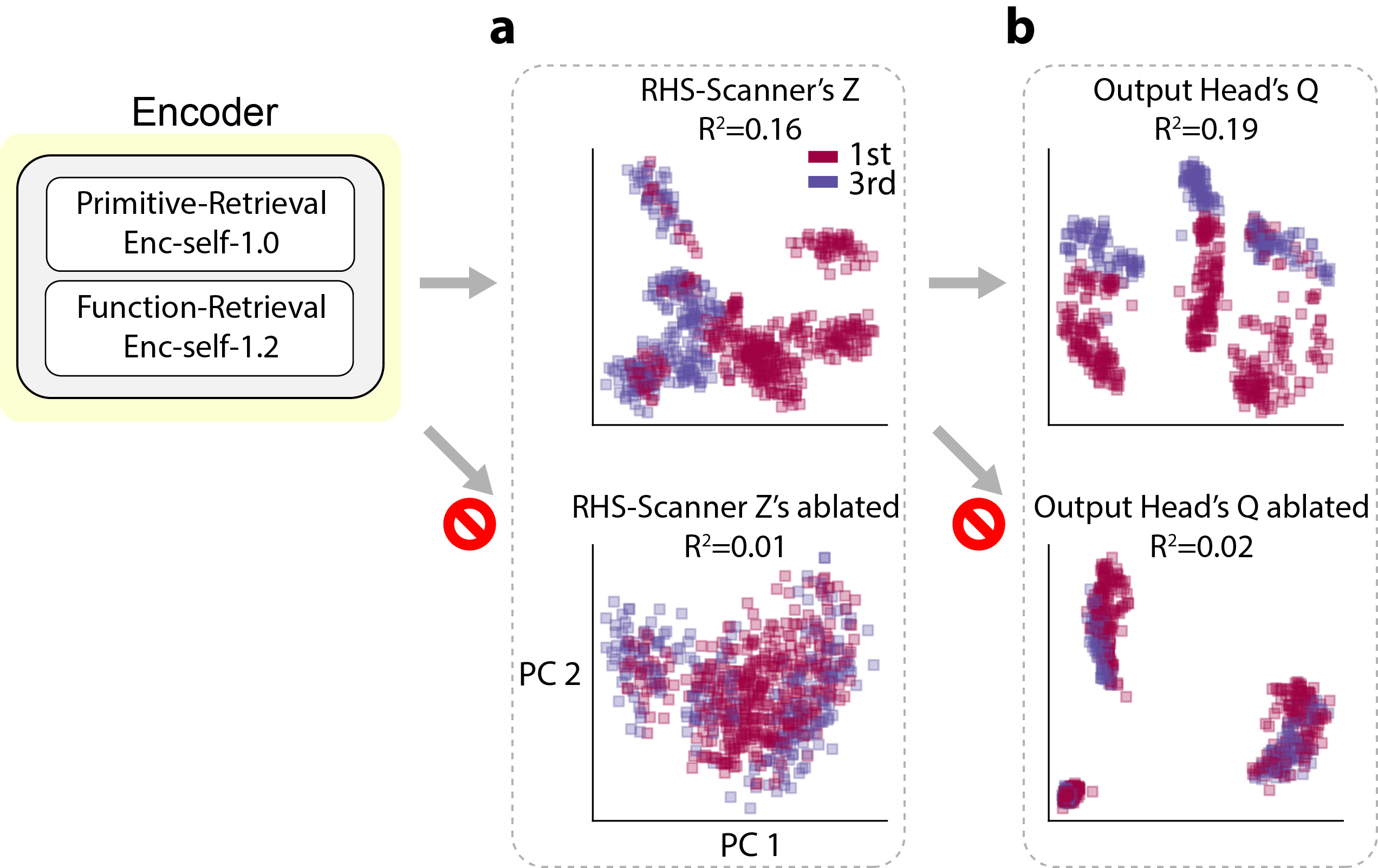}
\end{center}
\caption{PCA for token embeddings labeled by relative-index-on-LHS. Concretely, for an episode with prompt '\texttt{B S A | A=red | B=blue | A S B=blue red}' and prediction '\texttt{SOS red blue EOS}', in (a), points are the $Z$ of '\texttt{SOS}' and '\texttt{red}' in the decoder input tokens (\texttt{SOS} is labeled \texttt{3rd}, because \texttt{SOS} attends to the \texttt{blue} on function RHS, and \texttt{B} is the \texttt{3rd} on the LHS; similarly, \texttt{red} is labeled \texttt{1st}); in (b), points are the $Q$ of decoder input tokens  (\texttt{SOS} is labeled \texttt{3rd}, \texttt{red} is labeled \texttt{1st}).
 $R^2$ score quantifies the percentage of total variance explained by the index identity.}
\label{fig10}
\end{figure}

\paragraph{Relative-Index-On-LHS Tracing}
To confirm that our discovered circuit genuinely encodes the relative-index-on-LHS in the Output Head's $Q$, 
we conducted three complementary ablation experiments summarized in Figure \ref{fig10}:

\begin{itemize}

    \item \textbf{Retaining only the Primitive- and Function-Retrieval Heads} When all other encoder heads are 
    ablated, the RHS-Scanner Head's $Z$ still carries relative-index-on-LHS that propagate to the Output Head's $Q$, indicating that these two heads alone provide sufficient index information.

    \item \textbf{Ablating the Primitive- or Function-Retrieval Head individually} Ablating either head disrupts 
    the clustering by relative-index-on-LHS in the RHS-Scanner Head's $Z$, demonstrating that both heads are necessary to preserve the full index information.

    \item \textbf{Ablating the RHS-Scanner Head (together with Dec-cross-0.0 and -0.3)} These decoder heads share similar attention patterns that track color tokens on the function’s RHS. When all three are ablated, 
    clusterings by relative-index-on-LHS are eliminated from the Output Head's $Q$.
\end{itemize}

Thus, we conclude that the $Q$-circuit depends on the RHS-Scanner Head to capture the relative-index-on-LHS information supplied by the Primitive- and Function-Retrieval Heads. By aligning 
these $Q$ signals with the $K$, the model consistently determines which token to generate next.

\subsection{Targeted Perturbation Steers Behavior}
\label{section4.3}

So far, our circuit tracing indicates that the $K$-circuit of the Output Head encodes the primitive symbols' index-in-question, and that the $Q$-circuit encodes primitive symbols' relative-index-on-LHS. We reason that if the $QK$ circuit of the Output Head truly leverages on the \emph{primitive symbol index} to predict the next word, then \textbf{swapping} those index information across different color tokens should \emph{also} swap the corresponding attention
patterns observed in the Output Head.

\begin{figure}[h!]
\begin{center}
\includegraphics[width=0.32\textwidth]{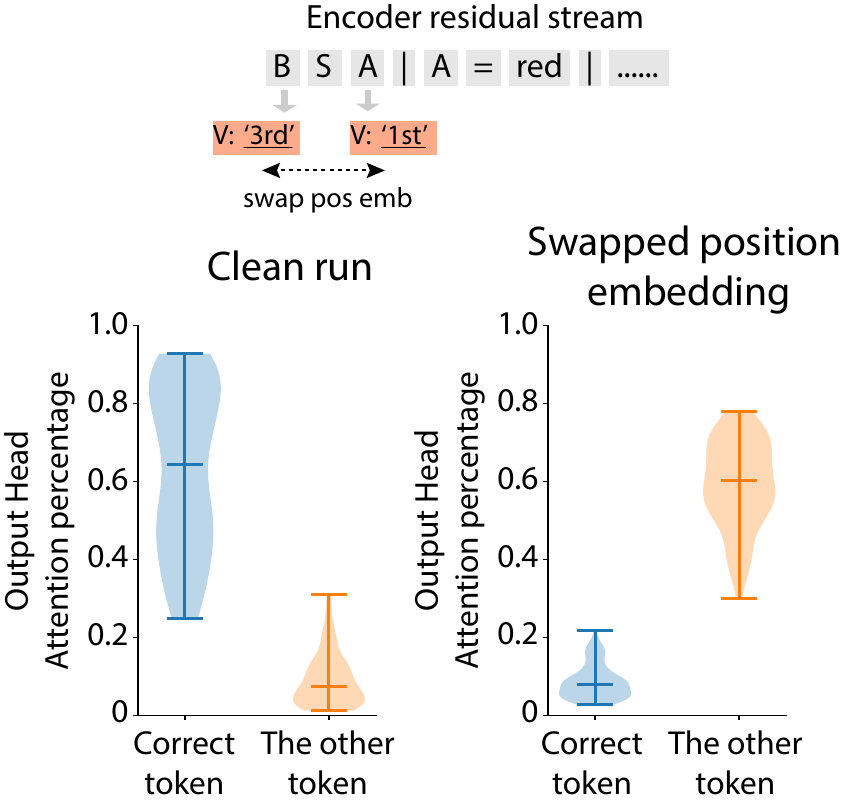}
\end{center}
\caption{Swapping position embeddings of tokens in the question causes a predictable realignment 
of attention in the Output Head through its $K$-circuit, confirming that the discovered $QK$ circuit indeed encodes positional indices.}
\label{fig11}
\end{figure}

\paragraph{Swapping Index Information}
Concretely, we select two primitive symbols in the question (e.g., '\texttt{\textbf{B} S \textbf{A} |} \texttt{A=red | B=blue |...}'). The \texttt{red} token will have index-in-quesiton=\texttt{3rd} from \texttt{A} (similarly \texttt{blue} will have '\texttt{1st}') on the $K$-side of the Output Head. If the $Q$-side expects a particular index from the $K$-side (e.g., '\texttt{SOS}' in $Q$ may carry relative-index-on-LHS=\texttt{3rd} and expects tokens carrying index-in-question=\texttt{3rd} from $K$), a swap of the index information in $K$ should lead to a predictable shift in which tokens the head attends to. We performed this perturbation in the $K$-circuit of the Output Head while \textbf{freezing} its $Q$-circuit.
Indeed, when we swap only the position embedding of \texttt{B} and \texttt{A} on the Question-Broadcast Head's $V$ (the most upstream node in the $K$-circuit), with everything else intact, we observe that the Output Head systematically “reverts” the attention from \texttt{red} to \texttt{blue} based on their swapped positions (Figure \ref{fig11}).

This intervention thus provides \emph{causal} evidence that the Output Head's $QK$ alignment relies
on the index information on both sides passed through the sub-circuits. It does not merely degrade or
randomly scramble the output head's behavior; rather, the predictions shift in a way directly consistent
with our interpretation of how index information is encoded and matched between $Q$ and $K$.
The model’s predictable response to this precise manipulation underscores that we have correctly
identified the sufficient pathways.

Overall, by performing causal backtracking, validating information flow through ablations, and
finally applying targeted activation patching, we confirm that the compositional induction mechanism we
uncovered is both \emph{interpretable} and \emph{causally relevant} to the model’s behavior.

\section{Discussion}

In this work, we investigated how a compact transformer model achieves compositional induction
on a synthetic function composition task. By combining \emph{path-patching} analyses with
\emph{causal ablations}, we uncovered a detailed $QK$ circuit that encodes index
information from both the question and the function's LHS. We further demonstrated
that precisely swapping these positional embeddings in the model’s activations leads to predictable
changes to behavior, thereby confirming the causal relevance of the discovered circuit. These results
show that, even for complex functions, transformers can implement a structured and
interpretable mechanism.

\subsection{Limitations and Future Work}

\textbf{Model Scale.} Our circuit analysis focused on a relatively small transformer. 
    Establishing whether similar interpretable circuits exist in larger models remains an important open question to follow up.
    \\
\noindent
\textbf{Manual Circuit Discovery.} The techniques employed here required substantial 
    human effort—path-patching, ablations, and extensive interpretation of attention heads. For 
    large-scale models, such manual approaches become less feasible. We therefore see a need 
    for automated or semi-automated methods that can discover and interpret these circuits 
    with less human input.

\noindent
\textbf{Partial Perturbations.} Although our targeted activations swaps successfully 
    steered the Output Head’s behavior, we have not demonstrated a complete perturbation of its predicted tokens. 
    This is due to the distributed nature of the underlying mechanism (multiple heads fulfill 
    similar roles). Coordinating interventions across all such heads will require systematic 
    workflows, which we aim to develop in the future.

Despite these constraints, our work shows that disassembling
transformer circuitry can yield \textit{two key benefits}. First, it illuminates how compositional
functions are mechanistically instantiated at the attention-head level. Second, it enables
targeted, activation-based interventions that reliably steer model behavior. We hope these
contributions will encourage further research on scalable circuit discovery methods and more
automated interpretability approaches for large-scale models.

\section{Acknowledgement}
CT was supported by the Friends of McGovern Fellowship.
MJ was supported by the Simons Foundation.

\nocite{ChalnickBillman1988a}
\nocite{Feigenbaum1963a}
\nocite{Hill1983a}
\nocite{OhlssonLangley1985a}
% \nocite{Lewis1978a}
\nocite{Matlock2001}
\nocite{NewellSimon1972a}
\nocite{ShragerLangley1990a}

\newpage

\bibliographystyle{main}

\bibliography{main}

\appendix
\section{Appendix}

\subsection{Transformer Model}
\label{appendix2}
We adopt an encoder-decoder architecture, which naturally fits the task by allowing the encoder to 
process the prompt (question + support) with bidirectional self-attention and the decoder to 
generate an output sequence with causal and cross-attention. Specific hyperparameters include:
\begin{itemize}
    \item Token embedding dimension: $d_{\mathrm{model}} = 128$
    \item Attention embedding dimension: $d_{\mathrm{head}} = 16$
    \item Eight attention heads per layer (both encoder and decoder)
    \item Pre-LayerNorm (applied to attention/MLP modules) plus an additional LayerNorm at the 
          encoder and decoder outputs
    \item Standard sinusoidal positional embeddings
\end{itemize}

The encoder comprises two layers of bidirectional self-attention + MLP, while the decoder comprises 
two layers of causal self-attention + cross-attention + MLP. We train the model by minimizing the 
cross-entropy loss (averaged over tokens) using the Adam optimizer. The learning rate is 
initialized at $0.001$ with a warm-up phase over the first epoch, then linearly decays to $0.00005$ 
over training. We apply dropout of $0.1$ to both input embeddings and internal Transformer layers, 
and train with a batch size of 25 episodes. All experiments are performed on an NVIDIA A100 GPU.

\subsection{Task Structure}
\label{appendix1}
In each episode, the \emph{support set} and \emph{question} are concatenated into a single prompt 
for the encoder, with question tokens placed at the start. Question, primitive assignments, and function assignments are separated by `\texttt{|}` 
tokens, while primitive and function assignments are identified by `\texttt{=}`. Overall, there are 
6 possible colors and 9 symbols that may serve as either color primitives or function symbols. 
Each episode contains 2--4 function assignments and 3--4 color assignments.  

A function may be a single-argument (\(\texttt{arg func}\)) or double-argument 
(\(\texttt{arg1 func arg2}\)) function. The function's right-hand side (RHS) describes how 
arguments are transformed, generated by randomly sampling up to length-5 sequences of arguments and 
mapping them to color tokens. Each prompt ends with an `\texttt{EOS}` token. During decoding, the 
model begins with an `\texttt{SOS}` token and iteratively appends each newly generated token until 
it emits `\texttt{EOS}`.

We randomly generate 10{,}000 episodes for training and 2{,}000 for testing, ensuring that the 
primitive and function assignments in testing episodes do not overlap with those in the training set.

\subsection{Path Patching}
\label{appendix3}

Path patching is a method for isolating how a specific \emph{source node} in the network influences a particular \emph{target node}. It proceeds in three runs:

\begin{enumerate}
    \item \textbf{Clean Run:} Feed the input through the model normally and \emph{cache} all intermediate 
    activations (including those of the source and target nodes).

    \item \textbf{Perturbed Run:} Freeze all direct paths into the target node using their cached 
    activations from the clean run. For the \emph{source node} alone, replace its cached activation 
    with \emph{mean-ablated} values. Record the new, perturbed activation at the target node.

    \item \textbf{Evaluation Run:} Supply the target node with the perturbed activation from 
    Step~2, then measure any resulting changes in the model’s output. This quantifies how the 
    source node’s contribution (altered via mean-ablation) affects the target node’s behavior.
\end{enumerate}

\paragraph{Chained Path Patching.}
When analyzing circuits that span multiple nodes in sequence, we extend path patching in a 
\emph{chain-like} manner. For instance, to evaluate a chain \(A \rightarrow B \rightarrow C\):
\begin{itemize}
    \item We first perform path patching on the sub-path \(B \rightarrow C\) as usual.
    \item Next, to capture how \(A\) specifically influences \(B\), we isolate and record 
    \(A\)’s effect on \(B\) via mean-ablation on all other inputs to \(B\).
    \item Finally, we patch that recorded activation into \(B\) and evaluate its effect on \(C\).
\end{itemize}
For a chain of length \(N\), we run \(N+1\) forward passes, ensuring the measured impact on the 
target node reflects only the chained pathway. This approach precisely attributes the model’s 
behavior to the intended sequence of dependencies.

\paragraph{Two Modes of Ablation.}
To assess how individual heads or nodes contribute to the target node, we use two complementary 
modes:
\begin{enumerate}
    \item \textbf{Keep-only-one-head:} Mean-ablate all direct paths to the target node except for 
    \emph{one} node, which retains its clean-run activation. If the target node’s performance 
    remains stable, this single node is \emph{sufficient} for driving the relevant behavior. 
    However, this method may fail when multiple heads each provide partial information that is 
    only collectively sufficient.

    \item \textbf{Ablate-only-one-head:} Keep all source nodes from the clean run except one, 
    which is mean-ablated. If performance degrades, that ablated node is \emph{necessary}. 
   However, if the node's information is \emph{redundant} or duplicated across other paths, the target node's performance will not significantly change.
\end{enumerate}

By combining both modes, we identified the putative QK-circuit of the output head. We then validate the circuits by inspecting the information they propagates and causally erasing the information by ablating specific upstream nodes.

\subsection{$R^2$ Score}
\label{appendix4}
To quantify how much an activation dataset $\mathbf{Y}$ encodes a particular latent variable 
$\mathbf{Z}$, we compute a linear regression of $\mathbf{Z}$ (one-hot encoded) onto $\mathbf{Y}$ 
and measure the explained variance:
\[
R^2 = 1 - \frac{SS_{res}}{SS_{total}}.
\]
An $R^2$ value of $1.0$ indicates that $\mathbf{Z}$ fully explains the variance in $\mathbf{Y}$, 
whereas an $R^2$ near $0.0$ implies $\mathbf{Z}$ provides no information about $\mathbf{Y}$.

\end{document}